\documentclass[letterpaper, 10 pt, conference]{ieeeconf}  % Comment this line out if you need a4paper

\usepackage{soul}
\usepackage{xcolor}
\usepackage{amsmath}
\usepackage{cite}
\usepackage{graphicx}
\usepackage{xcolor}
\usepackage{algorithm}
\usepackage{algpseudocode}
\usepackage{amsmath}
\usepackage{lipsum}
\usepackage{subcaption}
\usepackage{tabularx}
\usepackage{booktabs}
\usepackage{layouts} 
\usepackage{amsfonts}
\usepackage[utf8]{inputenc}

\IEEEoverridecommandlockouts                              % This command is only needed if 
\newtheorem{remark}{Remark}

\title{\LARGE \bf
Delay-Aware Diffusion Policy: Bridging the\\Observation–Execution Gap in Dynamic Tasks
}
\author{Aileen Liao$^{*}$, Dong-Ki Kim, Max Olan Smith, Ali-akbar Agha-mohammadi, Shayegan Omidshafiei\\\textbf{FieldAI}
\thanks{*AL conducted this work during their internship at FieldAI. AL is with the University of Pennsylvania.}}
\definecolor{navy}{RGB}{0,0,128}
\makeatletter
\let\NAT@parse\undefined
\makeatother
\usepackage[hidelinks,breaklinks=true]{hyperref}

\usepackage{capt-of}

\begin{document}

\maketitle

\thispagestyle{empty}
\pagestyle{empty}

%%%%%%%%%%%%%%%%%%%%%%%%%%%%%%%%%%%%%%%%%%%%%%%%%%%%%%%%%%%%%%%%%%%%%%%%%%%%%%%%
\definecolor{myorange}{HTML}{f19f78}
\sethlcolor{myorange}

\begin{abstract}
As a robot senses and selects actions, the world keeps changing. 
This inference delay creates a gap of tens to hundreds of milliseconds between the observed state and the state at execution.
In this work, we take the natural generalization from zero delay to measured delay during training and inference. We introduce Delay-Aware Diffusion Policy (DA-DP), a framework for explicitly incorporating inference delays into policy learning.  
DA-DP corrects zero-delay trajectories to their delay-compensated counterparts, and augments the policy with delay conditioning.
We empirically validate DA-DP on a variety of tasks, robots, and delays and find its success rate more robust to delay than delay-unaware methods.
DA-DP is architecture agnostic and transfers beyond diffusion policies, offering a general pattern for delay-aware imitation learning.
More broadly, DA-DP encourages evaluation protocols that report performance as a function of measured latency, not just task difficulty. Highlight videos can be found at: \textcolor{navy}{\url{https://dadpiros2026.github.io/}}.
\end{abstract}

%%%%%%%%%%%%%%%%%%%%%%%%%%%%%%%%%%%%%%%%%%%%%%%%%%%%%%%%%%%%%%%%%%%%%%%%%%%%%%%%
\section{Introduction}
Robotic control rarely happens in a static world. Sensors scan the environment while the world continues to change, so the state seen at exposure does not correspond to the state at actuation. 
We refer to the elapsed time between sensing and the moment the resulting command takes effect as the \emph{inference delay}. 
In real robotic systems, this delay can accumulate to tens to hundreds of milliseconds across sensing, networking, computation, and actuation. 
As a result, a policy that assumes zero delay will often act too late.

\begin{figure}[t]
    \centering
    \begin{subfigure}[b]{\columnwidth}
        \centering
        \includegraphics[width=0.98\linewidth,page=1]{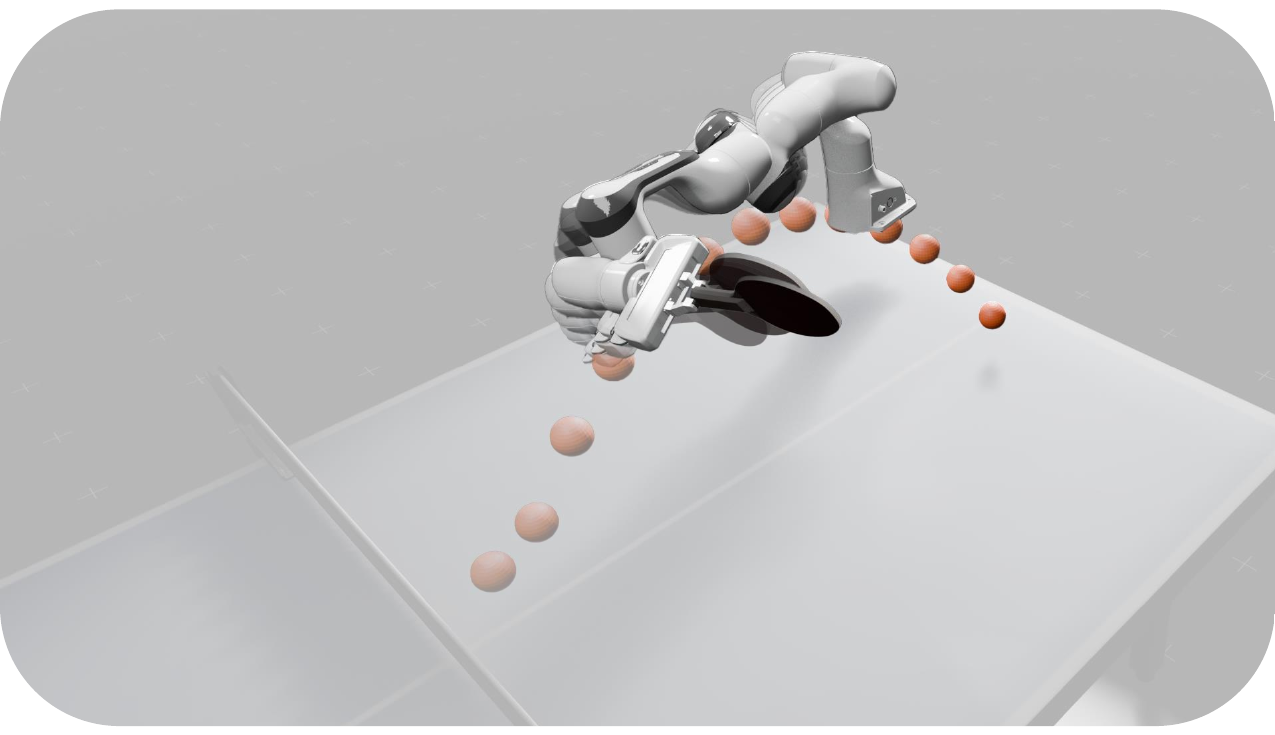}
        \vspace{-0.1cm}
        \caption{DP responds slowly to the moving ping-pong ball.}
        \label{fig:teaser-first}
    \end{subfigure}
    \hfill
    \begin{subfigure}[b]{\columnwidth}
        \centering
        \includegraphics[width=0.98\linewidth,page=2]{figures/pingpong_highlight.pdf}
        \vspace{-0.1cm}
        \caption{DA-DP (ours) is able to successfully hit the ping-pong ball.}
        \label{fig:teaser-second}
    \end{subfigure}
    \caption{While Diffusion Policy (DP) struggles with computation delays and fails to hit the ball in the ping-pong task, our Delay-Aware Diffusion Policy (DA-DP) successfully handles highly dynamic, reactive tasks under inference delays.}
    \label{fig:two_figs}
    \vspace{-0.3cm}
\end{figure}

This does not diminish how useful the zero inference delay assumption has been; it enabled clean supervision, stable training, and rapid progress across many tasks in a static world. 
Building upon these successes, the natural next step is to consider tasks with dynamic, high-speed interaction (e.g., returning a ping-pong ball), where inference delay becomes a first-order concern.
In these tasks, we now need to relax the zero-delay assumption in both data collection and algorithm design. 
This allows policies to plan for the state that will exist at execution time, not merely the one that was observed.

However, most data collection processes still assume zero inference delay--whether in teleoperation or simulation--for practical reasons (see simulation benchmarks~\cite{deepmind_control, robosuite, tao2025maniskill3} and real-world datasets~\cite{lynch2019play, ebert2021, brohan2023rt1roboticstransformerrealworld} without delay). In teleoperation, delays make it difficult for operators to anticipate and mimic behaviors when controlling robots. In simulation, incorporating variable delays complicates trajectory generation and can destabilize learning algorithms. 
Ignoring inference delays can be benign for static interaction. 
In dynamic environments, however, it opens a realism gap between observation and execution that can impair control.

Similarly, visuomotor policy learning approaches generally do not explicitly account for inference delay in their algorithmic design~\cite{chi2024diffusionpolicy, dasari2019robonet, james2019rlbench, yu2019metaworld}. 
In dynamic environments, the mismatch between the observed state and state at execution can lead to actions that always lag behind.
For example, consider a robotic arm needing to return a serve in ping-pong controlled via Diffusion Policy (DP)~\cite{chi2024diffusionpolicy}.
In ping-pong, inference delays can be depicted as in Fig.~\ref{fig:teaser-first} where the ball continues to move past the robot while it is computing an action.
So, an optimal policy must position the paddle where the ball will be at actuation, not where it was observed prior to inference. 
We find that DP struggles in this setting, as inference delay causes a systematic lag in responding to the moving ball (see Fig.~\ref{fig:teaser-first}).

In this paper, we introduce \textbf{Delay-Aware Diffusion Policy (DA-DP)}, a novel framework that improves diffusion policy performance in dynamic environments by explicitly modeling inference delay. 
Specifically, DA-DP first corrects training data collected under the zero-delay assumption by predicting delayed execution states and computing actions that properly transition between them. 
We then train a DP on the corrected data. 
We also condition the policy on measured delay so it accounts for changes between observation and execution.
Returning to our ping-pong example, DA-DP is able to account for the delays in inference and meet the ball in its new location as shown in Fig.~\ref{fig:teaser-second}.

In summary, our contributions are as follows:
\begin{enumerate}
  \item \textbf{DA-DP framework:} We propose an effective extension of DP that conditions action generation on the inference delay.
  \item \textbf{Empirical evaluation:} Through extensive experiments, we show that DA-DP significantly outperforms the DP baseline across a wide range of delay conditions, maintaining high success rates in challenging dynamic environments.
\end{enumerate}
Together, these contributions highlight inference delay as a critical yet underexplored challenge in visuomotor policy learning, and provide an effective solution that enhances robustness in dynamic robotic tasks.

\section{Related Work}
\textbf{Efficient DP.} A number of extensions have sought to improve efficiency and responsiveness. One-step distillation \cite{yin2024onestep} accelerates inference by collapsing the diffusion process into a single network pass, while Streaming Diffusion Policy~\cite{hoeg2024streamingdiffusionpolicyfast} produces partially denoised actions on-the-fly to reduce end-to-end latency. Consistency policy~\cite{prasad2024consistency} reduces the number of diffusion steps by enforcing self-consistency across different denoising stages.
These methods mitigate computational bottlenecks but still assume that the scene is static between sensing and execution. 

\textbf{Synchronous and asynchronous DP.} 
In synchronous DP~\cite{chi2024diffusionpolicy}, the policy has a zero-order hold while the next horizon of actions are being computed. 
In order to compensate the lack of actions during next horizon inference, asynchronous DP starts computing next horizon of actions while previous actions were still being executed~\cite{black2025realtimeexecutionactionchunking}. 
Overlapping horizon prediction and action execution makes the policy smoother and more continuous, but it introduces stale actions and requires marrying executed and predicted trajectories. 
For interaction with dynamic objects, both synchronous and asynchronous DP have observation-execution time mismatch. 
In other words, the world moves on, but the robot either waits (synchronous), or carries out stale actions (asynchronous) that may not align with the object anymore.
DA-DP is complementary to both paradigms: it augments datasets and conditions 
the policy on delays, making it plug-and-play regardless of the execution style.
Crucially, asynchronous DP~\cite{black2025realtimeexecutionactionchunking} and 
latency-matching approaches such as UMI~\cite{chi2024universal} operate at the 
execution level by overlapping inference with action playback, but they do not 
correct the fundamental observation-execution mismatch: the policy still plans 
relative to a stale observation, and in dynamic environments the world state 
continues to evolve during that overlap.
DA-DP instead addresses this at the \emph{data and training} level, explicitly 
conditioning the policy on $\delta$ so that generated actions target the 
execution-time state $s_{t+\delta}$ rather than the observation-time state $s_t$.
These two directions are therefore not interchangeable baselines but orthogonal 
contributions. Combining DA-DP's delay-aware training with asynchronous execution 
is a natural avenue for future work

\textbf{Model Based Control.} 
Classical approaches to delay handling often rely on Model Predictive Control (MPC), where trajectories are continuously replanned at execution time~\cite{MAYNE2000789}. 
MPC absorbs sensing and computation delays by forecasting forward from the current state and executing only the near-term portion of the plan. 
Many efforts~\cite{bhardwaj2021stormintegratedframeworkfast, nubertSafeMPC, mangalore2024neuromorphicquadraticprogrammingefficient,erez2013humanoids_mpc} aimed to make MPC faster to reduce latency. 

\textbf{Chunked and asynchronous execution.} 
Several approaches address real-time execution by planning in action chunks. 
Action Chunking Transformers~\cite{zhao2023learningfinegrainedbimanualmanipulation} generate smooth trajectories across discrete segments, while Real-Time Chunking (RTC)~\cite{black2025realtimeexecutionactionchunking} for flow policies overlaps inference with execution by freezing near-term actions and inpainting the remainder. 
Such methods still plan relative to exposure-time observations, without explicitly forecasting execution-time states.

\textbf{Reactive extensions.} 
An orthogonal line of work incorporates fast feedback via hierarchies of models, and incorporates dynamics more explicitly to correct actions during execution. Reactive Diffusion Policy~\cite{xue2025reactive} augments diffusion models with tactile or force signals, enabling fine-grained adaptation in contact-rich manipulation. 
This method relies on fast feedback loops from the lower level ``reactive'' controller, while the higher level DP plans next trajectories.

\textbf{Our approach.} 
DA-DP complements these efforts by treating inference delay as explicit design parameter. 
Instead of restructuring the denoising process or relying on additional sensing modalities, DA-DP conditions directly on forward-propagated execution-time states. 
This delay-aware supervision closes a critical gap in dynamic settings, enabling robust performance in fast, contact-rich tasks.

\begin{figure*}[!t]
    \centering
    \includegraphics[width=0.98\textwidth]{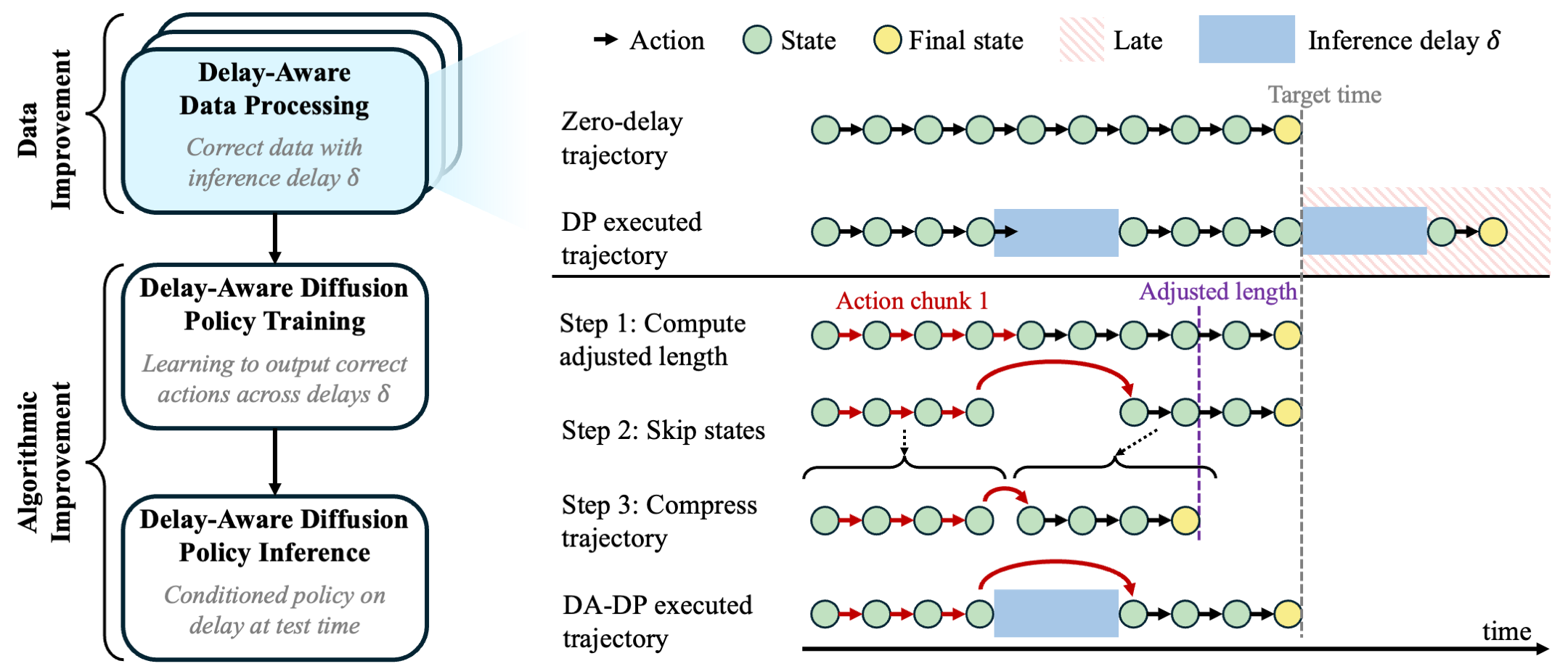}
    \caption{Overview of Delay-Aware Diffusion Policy (DA-DP). The DA-DP framework is depicted for the case with action chunk length $H_\text{act}=4$ and inference delay $\delta=2\Delta t$. DA-DP effectively modifies dataset trajectories by removing states and interpolating actions (e.g., Action chunk 1 in red) such that the trajectory reaches the final state at the ideal end time.}
    \label{fig:overview}
    \vspace{-0.3cm}

\end{figure*}

\section{Background}
\subsection{Diffusion Policy}
Diffusion policy (DP)~\cite{chi2024diffusionpolicy} adapts Denoising Diffusion Probabilistic Models (DDPMs)~\cite{ho2020ddpm} for sequential decision-making. 
In particular, an action sequence $\{a_{0},a_{1},\ldots\}$ is modeled as a sample from a generative process that gradually denoises Gaussian noise into a trajectory conditioned on the current state $s_t$. 
During training, noise is added to expert trajectories according to a forward process:
\begin{equation}
    q(a_t^{(k)} \mid a_t, k) = \mathcal{N}\!\left(\sqrt{\alpha_k}\, a_t,\,(1-\alpha_k)I\right),
\end{equation}
where $k$ denotes the diffusion timestep and $\alpha_k\in(0,1]$ is a noise scheduling coefficient that controls the signal-to-noise ratio at step $k$. 
Note that the forward process itself only corrupts the action $a_t$, not the state $s_t$. The conditioning on $s_t$ is introduced during the reverse process, where the policy learns to predict the clean action given both the noisy action and state. 
At inference, the policy iteratively samples from the reverse process to generate feasible actions that respect dynamics and task constraints. 
This formulation of DP enables the modeling of complex, multimodal action distributions while maintaining stability during training.

\subsection{Inference Delay Definition}
Inference delay refers to the temporal gap, denoted by $\delta$, between observing the environment state and executing the corresponding action. Our paper distinguishes between:
\begin{enumerate}
    \item \emph{Observation state $s_t$:} the state perceived at time $t$ 
    \item \emph{Execution state $s_{t+\delta}$:} the state when the corresponding action is executed by the robot after inference delay $\delta$.
\end{enumerate}
Such delays may arise from a combination of perception (e.g., sensing, image capture, pre-processing latency), computation (e.g., the cost of a model forward pass), and actuation (e.g., delays in low-level control, communication, or hardware execution). In real robotic systems, however, inference delay is inherently non-zero ($\delta\!>\!0$), leading to a systematic mismatch between states and actions, especially in highly dynamic environments (e.g., a ping-pong task). 
% In this work, we explicitly account for $\delta$ in both data-level and policy-level design.

\section{Delay-Aware Diffusion Policy}
Our work is motivated by the fact that inference delays inherent to real robotic systems create a gap between observation and execution, leading to actions that lag behind the current environment state in dynamic settings.
% slower actions in dynamic environments. 
This section first introduces our objective of DA-DP that explicitly models inference delay (Section~\ref{sec:da-dp-objective}). We then describe how DA-DP improves DP at both the training-data level (Section~\ref{sec:delay-aware-data-processing}) and the algorithmic level (Section~\ref{sec:da-dp-policy}). An overview of the DA-DP framework is provided in Fig.~\ref{fig:overview}.

\subsection{Objective of DA-DP}\label{sec:da-dp-objective}
Formally, a trajectory $\tau\!=\!\{s_0, a_0, \ldots, s_n\}$ in an imitation learning dataset with zero delay transitions the initial state $s_0$ to the final state $s_n$ over a duration of:
\begin{equation}
T_\text{target} = n\Delta t,
\end{equation}
where $\Delta t$ denotes the control timestep (see Fig.~\ref{fig:overview}; \textit{Zero-delay trajectory}). For dynamic tasks such as hitting a ping-pong ball, the robot must reach $s_n$ precisely at $T_\text{target}$, as any delay means the ball will already be gone.

In practice, diffusion policies require inference time in computing actions. Specifically, every chunk of $H_{\text{act}}$ actions requires an additional inference delay $\delta$. This means the total inference time becomes:
\begin{equation}
T_\text{dp} = n\Delta t + (\lceil\frac{n}{H_{\textit{act}}}\rceil - 1)\delta.
\end{equation}
Because $T_\text{dp} > T_\text{target}$, the robot’s actions always lag behind the demonstration. In practice, this systematic delay can be fatal for dynamic tasks. For the ping-pong task, the robot arrives at the final state $s_n$ at time $T_\text{dp}$—too late to hit the ball (see Fig.~\ref{fig:overview}; \textit{DP executed trajectory}).

Our key objective is to adjust the training data so that the resulting trained policies can still act on time. Specifically, we construct a shorter trajectory $\tau'_\delta = \{s'_0, a'_0, \ldots, s'_{n'}\}$ that still ends at the same final state $s_n$ (i.e., $s'_{n'}=s_n$), but within the target duration $T_\text{target}$ for a given $\delta$:
\begin{equation}\label{eqn:da-dp-objective}
n'\Delta t + (\frac{n'}{H_{\textit{act}}} - 1)\delta = T_{\text{target}}.
\end{equation}
By explicitly accounting for inference delay $\delta$, this corrected trajectory $\tau'$ ensures that DP trained on it will reach $s_n$ right on schedule (see Fig.~\ref{fig:overview}, \textit{DA-DP executed trajectory}).

\subsection{Delay-Aware Data Processing}\label{sec:delay-aware-data-processing}
We detail how DA-DP constructs the delay-aware trajectory $\tau'$ in three steps.

\textbf{Step 1: Compute adjusted length.}
Solving Equation~\ref{eqn:da-dp-objective}, the corrected trajectory length $n'$ is given by:
\begin{equation}
    n' = \frac{n\,\Delta t + \delta}{\Delta t + \delta/H_{act}},
    \label{eqn:n'}
\end{equation}
where Fig.~\ref{fig:overview} (\textit{Step 1}) illustrates the case with $H_\text{act}=4$ and $\delta=2\Delta t$ as an example. 

\begin{remark}[Discrete implementation]\label{remark:1}
The expression in Equation~\ref{eqn:n'} defines the effective trajectory length $n'$ in continuous time. 
In practice, $n'$ must be an integer horizon since trajectories are discrete. 
We therefore select an integer $N' \in \{\lfloor n' \rfloor, \lceil n' \rceil\}$ that best satisfies the time constraint, 
and distribute the skipped steps across inference blocks accordingly. 
This ensures the compressed trajectory terminates within the original zero-delay duration. 
If $\delta/\Delta t$ is non-integer, the per-chunk skip is rounded and any residual mismatch is corrected in the final block.
\end{remark}

\begin{algorithm}[h]
\caption{Delay-Aware Data Processing}
\label{alg:delay-aware-data-processing}
\begin{algorithmic}[1]
\Require Zero-delay trajectory $\tau$, control timestep $\Delta t$, inference delay $\delta$, action chunk horizon $H_{\text{act}}$
    \State Compute zero-delay trajectory length $n$
    \Statex \textcolor{gray}{\textit{\# Step 1: Compute adjusted length}}
    \State Compute adjusted length $n'$ using Equation~\ref{eqn:n'}
    \State Initialize $\tau'=\{s'_0,a'_0,\ldots,s'_{n'}\}$ with $s'_{n'}=s_n$
    \Statex \textcolor{gray}{\textit{\# Step 2: Skip states}}
    \State Compute skip amount $m$ using Equation~\ref{eqn:shift}
    \Statex \textcolor{gray}{\textit{\# Step 3: Compress trajectory}}
    \For{$i = 0$ to $n'-1$} 
        \State Compute index $j=i+k(i)m$ using Equation~\ref{eqn:indexing}
        \State Update state $s'_i\in\tau_\delta'\gets s_j\in\tau$
    \EndFor
    
    \Statex \textcolor{gray}{\textit{\# Construct actions}}
    \For{$i = 0$ to $n'-1$} 
        \State Update action $a'_{i}\in\tau'_\delta \gets s'_{i+1} - s'_i$
    \EndFor
    \State \textbf{Return} Delay-aware trajectory $\tau_\delta$
\end{algorithmic}
\end{algorithm}

\textbf{Step 2: Skip states.} 
Next, we determine how many states to skip for each $H_{\text{act}}$ action chunk so that the compressed trajectory has the correct length $n'$ and still reaches the final state $s_n$ on time. Specifically, the skip amount $m$ is:
\begin{equation}
m =
\begin{cases}
\frac{\delta}{\Delta t}, & \text{for continuous time} \\
(n-n') /( \lceil \frac{n'}{H_{act}} \rceil-1), & \text{for discrete time}
\end{cases}
\label{eqn:shift}
\end{equation}
where Fig.~\ref{fig:overview} (\textit{Step 2}) shows the $m=2$ case.
Note that the discrete $m$ may differ from $\delta/\Delta t$ due to 
rounding; residual mismatches are absorbed in the final block (Remark~\ref{remark:1}).

% \textcolor{blue}{\noindent\textbf

\begin{remark}[Skip direction and interpolation]
Skipping $m$ states at the beginning versus end of each chunk is equivalent up to a global index shift: both remove the same states and preserve $s_n$ as the terminal state.
We adopt the beginning-skip convention as it directly aligns each chunk's initial state with the robot's execution-time state after delay $\delta$.
When $\delta/\Delta t \notin \mathbb{Z}$, the skip index falls between recorded states; we resolve this via linear interpolation between the two nearest neighbors in $\tau$.\end{remark}

\textbf{Step 3: Compress trajectory.} We construct the compressed state sequence $\{s'_0, a'_0, \ldots, s'_{n'}\}$ of $\tau'$ by skipping $m$ states after every chunk of $H_{\text{act}}$ actions. For each index $i \in \{0, \ldots, n'-1\}$, we define the mapping as:
\begin{equation}
    s'_i = s_{j}
    \qquad \text{ with }
    j = i + k(i)\,m,
    \label{eqn:indexing}
\end{equation}
where $k(i)=\lfloor i/H_{act} \rfloor$ counts how many action chunks have been completed up to $i$. This mapping results in the sequence: 
\begin{equation}
\{\underbrace{s_0, \ldots, s_{H_{\text{act}}-1}}_{\text{From first chunk}}, \underbrace{s_{H_{\text{act}}+m},\ldots,s_{2H_{\text{act}}+m-1}}_{\text{From second chunk}}, \ldots, s_n\}.
\end{equation}
In other words, after each chunk of actions, we jump ahead by $m$ states to offset inference delay (see Fig.~\ref{fig:overview}; \textit{Step 3}). Depending on the task, we then apply optional smoothing between states to ensure more natural robot behavior. Finally, we compute corresponding action sequences of $\tau'$ that transition between compressed states as $a'_i = s'_{i+1} - s'_{i}$ for $i\in\{0,\ldots,n'-1\}$.

\subsection{Diffusion Policy with Inference Delays}\label{sec:da-dp-policy}
To make our DA-DP policy robust to different inference delays, we create a set of delay-aware training datasets, $\{\tau'_\delta\}$, by varying the delay parameter $\delta$ (see Section~\ref{sec:delay-aware-data-processing}). We then combine these datasets to jointly train DA-DP, allowing the policy to learn not only how to predict actions from states but also how to adapt to different delays. This is achieved by explicitly conditioning the policy on the delay, denoted as $\pi_{\theta}(a'|s', \delta)$. Compared to DP, the main algorithmic difference is this explicit conditioning on $\delta$ during training and testing, which makes DA-DP easy to integrate into the DP framework while still highly effective.

We note that the inference delay $\delta$ of a robotic system can be easily measured by running one or a few control cycles and recording the elapsed time between sensing and action execution. If a system latency range is already available from hardware specifications, it could also be provided directly as an input condition to DA-DP.
In practice, DA-DP does not assume a fixed control timestep $\Delta t$; by training across 
a distribution of delay values $\delta$, the policy implicitly covers the 
variability in $\Delta t$ that arises from timing jitter in real hardware. 
At inference, the measured $\delta$ can be updated each control cycle to 
reflect the current system latency, naturally accommodating fluctuations 
without architectural changes.

\begin{algorithm}
\caption{DA-DP Training and Inference}
\label{alg:dp-train-and-inference}
\begin{algorithmic}[1]
\Statex \hspace*{-1.8em}\textbf{Training}
\Require A set of delay-aware trajectories $\{\tau_\delta\}$
\State Initialize diffusion policy $\pi_\theta$
\While{not converged}
    \State Sample batch $(\mathbf{s}, \mathbf{a}, \delta)\sim\{\tau_\delta\}$
    \Statex \hspace{1.5em}\textcolor{gray}{\textit{\# Augment $\delta$ into conditions for DP}}
    \State \hl{Form augmented state $\tilde{\mathbf{s}} \gets \operatorname{concat}(\mathbf{s}, \delta)$}
    \State Sample noise $\epsilon$ and diffusion timestep $t$
    \State Forward diffusion $\tilde{\mathbf{a}}_t \gets q(\mathbf{a}, \epsilon, t)$
    \State $\hat{\epsilon} \gets \pi_\theta(\tilde{\mathbf{a}}_t, t \mid \tilde{\mathbf{s}})$
    \State $\mathcal{L} \gets \|\hat{\epsilon} - \epsilon\|^2$
    \State Update $\theta \gets \theta - \eta \nabla_\theta \mathcal{L}$
\EndWhile
\State \textbf{Return} Trained delay-aware diffusion policy $\pi_\theta$
\Statex \hspace*{-1.8em}\textbf{Inference}
\Require Delay-aware diffusion policy $\pi_\theta$, inference delay $\delta$, environment $\mathcal{E}$, task horizon $T$
\State Reset environment $\mathcal{E}$
\For{$t=1$ to $T$}
    \State Get current state from environment $s\gets\mathcal{E}$
    \State \textcolor{gray}{\textit{\# Augment $\delta$ into conditions for DP}}
    \State \hl{Form augmented state $\tilde{s} \gets \operatorname{concat}(s, \delta)$}
    \State Sample delay-aware action $a\sim\pi_\theta(\cdot|\tilde{s})$ 
    \State Execute delay-aware action $a$ in the environment $\mathcal{E}$
\EndFor
\end{algorithmic}

\end{algorithm}
\subsection{DA-DP Algorithm}\label{sec:algorithm}
To clearly describe DA-DP, we first present the delay-aware data processing procedure in Algorithm~\ref{alg:delay-aware-data-processing}, followed by the training and inference algorithm for DA-DP in Algorithm~\ref{alg:dp-train-and-inference}. Minimal changes (highlighted in orange) are made to the baseline diffusion policy framework, making our approach easy to adapt to other policy architectures.

\begin{figure*}[!t]
    \centering
    \includegraphics[width=0.99\textwidth]{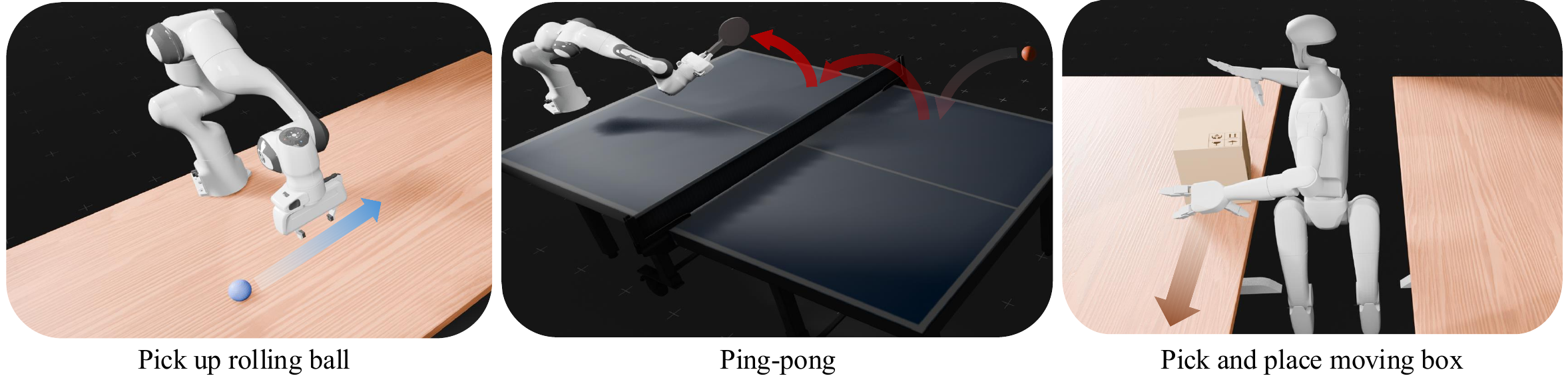}
    \caption{Three dynamic environments for experimental evaluations. The pick up rolling ball task consists of a ball rolling across a table and the Franka Emika Panda robotic arm picking it up and holding it at a specified location. The ping-pong task included the Panda arm hitting a ball after a serve. Lastly, the Pick and place moving box task included a G1 Unitree Humanoid picking up a box sliding across the table and placing it on the opposing table.}
    \label{fig:domain-overview}
    \vspace{-0.4cm}
\end{figure*}

\section{Experiments}
In this section, we empirically evaluate DA-DP across diverse tasks, robots, and delay conditions. 
Our experiments are designed to answer five guiding questions: 

\noindent \textbf{Q1.} Does inference delay impact performance on dynamic manipulation tasks? \\
\textbf{Q2.} Can DA-DP handle varying inference delays?\\
\textbf{Q3.} Is DA-DP robust to out of training distribution delays? \\
\textbf{Q4.} Does DA-DP scale to higher-dimensional  embodiments?\\
\textbf{Q5.} Does simulation trained DA-DP remain physically executable on the actual robot?
%generalize across different robot morphologies? 

These evaluations provide a comprehensive assessment of DA-DP’s robustness compared to delay-unaware baselines.

\subsection{Experiment setting}
\textbf{Domains.} We demonstrate DA-DP's effectiveness on three dynamic domains (see Fig.~\ref{fig:domain-overview}), implemented in ManiSkill~\cite{tao2025maniskill3}.
\begin{enumerate}
  \item\textbf{Pick up rolling ball:} The objective of this task is for the Franka Emika Panda arm to pick up a rolling ball from a table and hold it at a specified goal location. The ball location and velocity are randomized within a fixed range. The task uses an incremental Cartesian end-effector controller. %delta end-effector controller. 
  The robot action consists of $x$, $y$, $z$ of the end-effector position and a gripper position.
  \item\textbf{Ping-pong:} We design a custom table tennis environment in which a Franka Emika Panda arm strikes a ball so that it bounces once on its side before crossing the net and landing on the opponent’s side. The ball is initialized at a fixed position, and the paddle starts at the tool-center point. An external force initiates the ball’s motion. A proportional-derivative (PD) joint-delta controller is used with an 8-dimensional action space (7 joint positions and 1 gripper).
  % \item\textbf{Ping-pong:} We design a custom table tennis environment, where the Franka Emika Panda arm must hit a ball so that it bounces once on its own side before crossing the net and landing on the opponent’s side. The ball is placed at a fixed location in front of the robot. The paddle is initialized at the robot’s tool-center point pose. An external force is applied to the ball to induce motion at the initial time. A proportional-derivative (PD) joint-delta controller
  % %delta joint controller 
  % is used with an action space of $8$ (i.e., 7 joint positions + 1 gripper).
  \item\textbf{Pick and place moving box:} We design a custom box transfer environment in which a Unitree G1 humanoid picks up a moving box from one table and places it on an opposing table. The box is initialized at a random position with an initial velocity of 2.5 m/s. Control uses a proportional-derivative (PD) joint-delta controller with a 25-dimensional action space.
  
  % We design a custom box transfer environment in which a Unitree G1 humanoid picks up a moving box from one table and places it on an opposing table. The box is initialized at a randomized position and moves with an initial velocity of 2.5 m/s. Control is achieved using a proportional-derivative (PD) joint-delta controller with an action space of 25 joints.
\end{enumerate}

\textbf{Baselines.} We compare DA-DP against the followings:
\begin{enumerate}
\item \textbf{Diffusion policy~\cite{chi2024diffusionpolicy}:} We include the standard diffusion policy as a baseline. DP models actions as a conditional denoising diffusion process, where trajectories are iteratively refined from noise given the current observation. At test time, the policy generates a sequence of future actions and executes the first step in the environment. This baseline represents the state of the art in imitation learning for robotic manipulation, but does not account for inference delay.
\item \textbf{Zero-delay DP:} This baseline shows the performance of DP in an idealized setting with zero inference delay.
\end{enumerate}

\textbf{Implementation details.} We implement DA-DP in PyTorch~\cite{paszke2019pytorch} and evaluate on 100 environments. Task configurations are listed below:
\begin{enumerate}
\item \textbf{Pick up rolling ball:} Data is collected using motion planning, with 100 demonstrations. Models are trained for 30,000 iterations using AdamW~\cite{adamw} with an initial learning rate of 1e-4, a cosine decay schedule with 500 warmup steps, and a batch size of 256. The policy network is a 1D UNet with channel dimensions $[64, 128, 256]$. The observation horizon is set to 2, and the action horizon to 8.
\item \textbf{Ping-pong:} Data is collected via imitation learning from reinforcement learning agents trained with Proximal Policy Optimization ~\cite{DBLP:journals/corr/SchulmanWDRK17}. Models are trained with a batch size of 512, using 300 demonstrations, for up to 60,000 iterations. The policy network is a UNet with dimensions $[128, 256, 512]$.
\item \textbf{Pick and place moving box:} Data is collected using reinforcement learning. Models are trained for up to 150,000 iterations with a batch size of 512 and 200 demonstrations. Training uses a fixed learning rate of 5e-4, 1,000 denoising steps, and no scheduler. The network is a UNet with dimensions $[256, 512, 1024]$.
\end{enumerate}

\subsection{Main experiments}
Across all experiments, we set inference delays $\delta$ to reflect real-time latency. 
Prior work by~\cite{chi2024diffusionpolicy} reports an average inference delay of approximately 0.1s in real-world tests of diffusion policies. 
% Using this as a reference, we select a range of practical delays that align with the dynamics of each environment.
In practice, delays may vary under different computational loads; 
we therefore train across a range of $\delta$ values rather than a single fixed 
delay, ensuring the policy remains robust to this variability at test time.

\begin{figure}[h]
  \centering
  \includegraphics[width=\columnwidth]{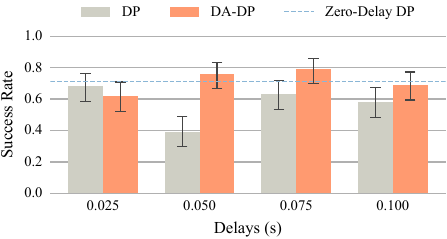}
    \caption{Performance comparisons in the ping-pong domain across different constant inference delays.
    }
  \label{fig:pingpong:delayed}
\end{figure}

\textbf{Q1.} \emph{Does inference delay impact performance on dynamic manipulation tasks?}

\begin{figure}[t]
  \centering
  \includegraphics[width=\columnwidth]{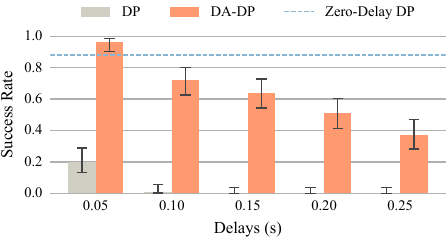}
   \caption{Performance comparisons in the pick up rolling ball domain across different constant inference delays.}
  \label{fig:placesphere:delayed}
      \vspace{-0.3cm}

\end{figure}

We evaluate both DP and DA-DP on datasets where a constant inference delay is applied. 
Across tasks, we observe that inference delay significantly degrades the performance of DP, whereas DA-DP maintains robustness.
In the pick up rolling ball task, DA-DP achieves success rates of 0.96 and 0.72 at $\delta = 0.05$s and $\delta = 0.10$s, compared to DP’s success rates of 0.20 and 0.01. Notably, as inference delays increase further, DP’s performance rapidly collapses to zero, while DA-DP degrades more gradually (see Fig.~\ref{fig:placesphere:delayed}). The marginal outperformance at $\delta=0.05$s is within expected stochastic variation from diffusion policy sampling.
In the ping-pong task, DP and DA-DP perform comparably under the smallest delay of $\delta = 0.025$s (see Fig.~\ref{fig:pingpong:delayed}). 
However, in all other larger delay settings, DA-DP consistently outperforms DP and even sustains performance close to the zero-delay DP baseline.

\begin{figure}[h]
  \centering
  \includegraphics[width=\columnwidth]{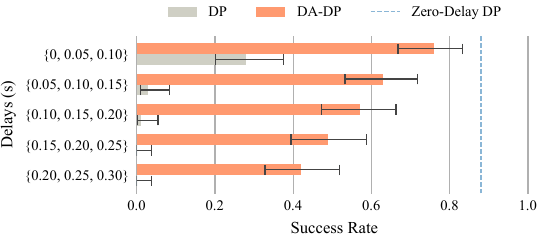}
  \caption{Performance comparisons in the pick up rolling ball domain across different sets of inference delays.
  }
  \label{fig:placesphere:delay-sets}
\end{figure}

\textbf{Q2.} \emph{Can DA-DP handle varying inference delays?}

\begin{figure}[h]
  \centering
  \includegraphics[width=\columnwidth]{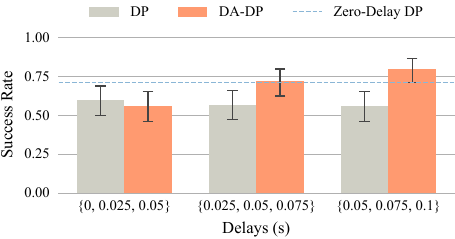}
  \caption{Performance comparisons in the ping-pong domain across different sets of inference delays.
  }
  \label{fig:pingpong:delay-sets}
\end{figure}

In this experiment, we train each method on a dataset consisting of multiple inference delay values. This setup more closely reflects real-world conditions, where inference delays are not fixed but instead vary within a range. Fig.~\ref{fig:placesphere:delay-sets} shows results for the pick up rolling ball task. Across all delay sets, DA-DP consistently outperforms DP. As delay values increase, DA-DP maintains a success rate between 0.76 and 0.42, while DP achieves only 0.28 even under the lowest-delay case. For the ping-pong task (see Fig.~\ref{fig:pingpong:delay-sets}), DP and DA-DP achieve similar success rates in the low-delay set ($\delta = \{0, 0.025, 0.05\}$). However, as delay values increase, DA-DP’s success rate improves, reaching 0.80. In contrast, DP’s performance remains largely unchanged across the different delay sets. The improving performance with larger delay sets may be attributed 
to discretization: for small $\delta$, rounding in the discrete skip amount $m$ can result in fewer states being dropped than prescribed, reducing the effectiveness of the delay correction. Larger $\delta$ values produce more pronounced skips that better reflect the true delay offset. These results suggest that training DA-DP with varying inference delays makes our policy more robust to dynamic delays at test time compared to DP.

\begin{figure}[h]
  \centering
  \includegraphics[width=\columnwidth]{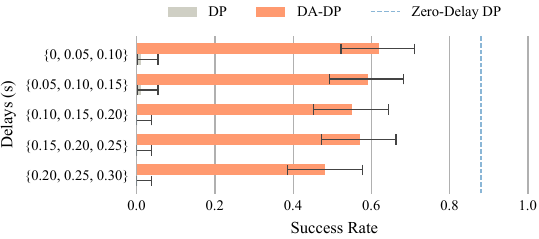}
  \caption{Pick up rolling ball: out of distribution delays.
  All methods were trained on a dataset composed of the labeled delays, and evaluated on a dataset of the original set +0.15s.
  }
  \label{fig:placesphere:ood}
\end{figure}

\textbf{Q3.} \emph{Is DA-DP robust to out of training distribution delays?}

In this experiment, we train methods on a fixed set of delays, and then evaluate them on a different set of delays.
We construct the out-of-distribution inference delay set by shifting the training set by constant factors depending on environment dynamics.
In our experiments, we find that DA-DP better generalizes to new delays than the baseline DP.

\begin{figure}[h]
  \centering
  \includegraphics[width=\columnwidth]{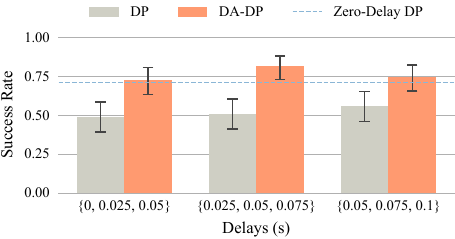}
  \caption{Ping-pong: out of distribution delays.
    All methods were trained on a dataset composed of the labeled delays, and evaluated on a dataset of the original set +0.075s.
  }
  \label{fig:pingpong:ood}
\end{figure}

In the pick up rolling ball environment the methods were evaluated on datasets with an increase of 0.15s to the training inference-delay set (see Fig.~\ref{fig:placesphere:ood}).
DP has near-zero performance across all delay sets.
In comparison, DA-DP maintains a performance rates between 0.62 and 0.48 for all delay sets.
In ping-pong, the methods were instead evaluated on delays increased by 0.075s (see Fig.~\ref{fig:pingpong:ood}).
DA-DP had a consistent performance between 0.73 and 0.82 across all delay sets, whereas DP performed between 0.49 and 0.56.

\begin{figure}[h]
  \centering
  \includegraphics[width=\columnwidth]{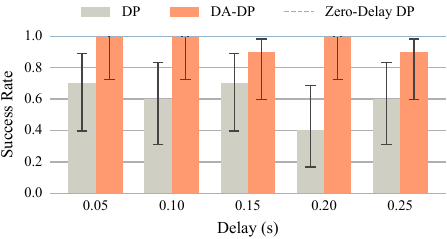}
  \caption{Performance comparisons in the pick and place moving box domain across constant inference delay.}
  \label{fig:humanoid:delayed}
      \vspace{-0.4cm}

\end{figure}

\textbf{Q4.} \emph{Does DA-DP scale to higher-dim embodiments}?

In this experiment, we evaluate whether DA-DP scales to higher-dimensional robot embodiments.
The previous experiments have centered around the Franka Emika Panda arm.
We now investigate whether the previous trends extend to the Unitree G1 Humanoid.
The Panda arm has 8 degrees of freedom (DOF) including the gripper, and the humanoid has 25 DOF, making the task more challenging. 
We repeat the same experimental methodology as in \textbf{Q1}, but now on the pick and placing moving box environment.
Fig.~\ref{fig:humanoid:delayed} shows our results.
In our experiment, we found that DA-DP was able to maintain a perfect success rate in three delay cases ($\delta=0.05\text{s}, 0.10\text{s}, 0.20\text{s}$); whereas, DP in the same cases performed at best 0.7.
In the other delay cases unilaterally DA-DP outperforms DP.
This experiment suggests that DA-DP is indeed able to scale to higher-dimensional robots.

\begin{figure}[t]
  \centering
  \includegraphics[width=\columnwidth]{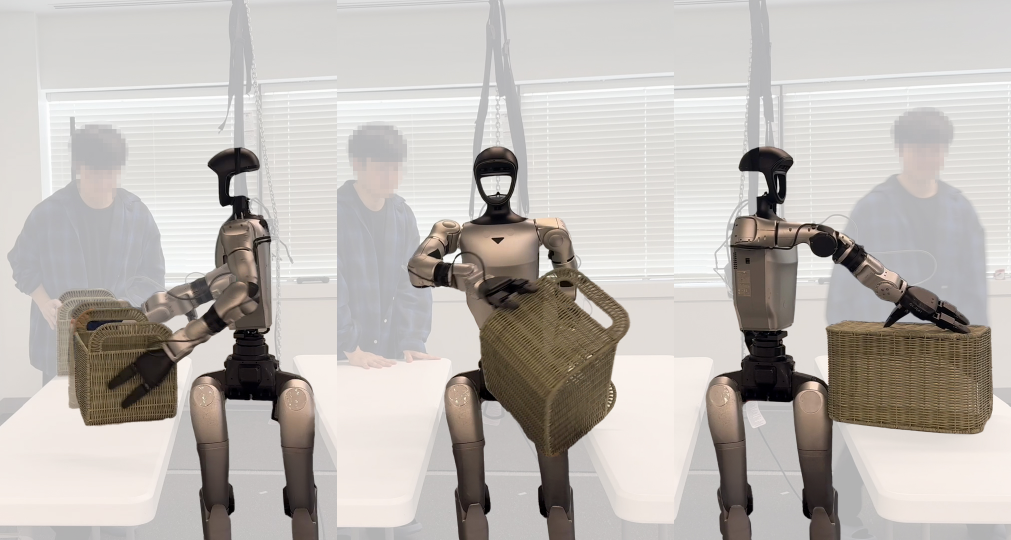}
  \caption{Timelapse of real-world execution. Frames show task progression from left to right.}
  \label{fig:hardware_timelapse_placeholder}
    \vspace{-0.4cm}

\end{figure}

\textbf{Q5.} \emph{Does simulation trained DA-DP remain physically executable on the actual robot?}

While skipping states may appear to introduce discontinuities between chunks, the simulation rollouts demonstrate that compressed trajectories remain physically executable. Furthermore, the optional smoothing step and linear interpolation  (for non-integer $m$) ensure that consecutive states are reachable within the  robot's kinematic limits. We utilized G1 hardware (Figure ~\ref{fig:hardware_timelapse_placeholder}) to execute the DA-DP trajectory learned in simulation under an inference delay of 0.1s. Our trajectory replay confirms that the compressed waypoints lie within the robot's reachable workspace.

% \begin{figure}[t]
%     \centering
% 
%     \begin{subfigure}[t]{0.18\linewidth}
%         \centering
%         \fbox{\parbox[c][3cm][c]{\linewidth}{\centering $t_0$}}
%         \caption{Start}
%     \end{subfigure}
%     \hfill
%     \begin{subfigure}[t]{0.18\linewidth}
%         \centering
%         \fbox{\parbox[c][3cm][c]{\linewidth}{\centering $t_1$}}
%         \caption{Early}
%     \end{subfigure}
%     \hfill
%     \begin{subfigure}[t]{0.18\linewidth}
%         \centering
%         \fbox{\parbox[c][3cm][c]{\linewidth}{\centering $t_2$}}
%         \caption{Mid}
%     \end{subfigure}
%     \hfill
%     \begin{subfigure}[t]{0.18\linewidth}
%         \centering
%         \fbox{\parbox[c][3cm][c]{\linewidth}{\centering $t_3$}}
%         \caption{Late}
%     \end{subfigure}
%     \hfill
%     \begin{subfigure}[t]{0.18\linewidth}
%         \centering
%         \fbox{\parbox[c][3cm][c]{\linewidth}{\centering $t_4$}}
%         \caption{End}
%     \end{subfigure}
% 
%     \caption{\textcolor{blue}{Timelapse of real-world execution (placeholder).
%     Frames show task progression from left to right.}}
%     \label{fig:hardware_timelapse_placeholder}
% \end{figure}
\section{Conclusion}
DA-DP is a predictive diffusion policy capable of handling dynamic objects and environments. 
Our method provides a principled approach to closing the observation–execution gap and scales to higher-dimensional embodiments, tasks, and inference delays. Additionally, DA-DP shows improved performance to larger out-of-distribution delays compared to standard diffusion policy. This work can naturally extend to other predictable, systematic delays in the control loop, offering a framework for more robust, responsive control.

\textbf{Future work.} 
While asynchronous DP reduces end-to-end latency by streaming partially denoised actions, it still suffers from stale predictions in dynamic conditions. DA-DP’s delay conditioning could complement asynchronous execution by explicitly training the policy to anticipate these residual delays, improving robustness when streamed trajectories lag behind the real state. More generally, the same principle applies beyond robotics: any domain with systematic, predictable delays, such as networked control, autonomous driving, or interactive simulation, could benefit from delay-aware conditioning to increase resilience to inference delays.

\nocite{*}
\bibliographystyle{IEEEtran}
\bibliography{refs}

\end{document}